%% file: main.tex
\documentclass[letterpaper, 10 pt, conference]{ieeeconf}  

\IEEEoverridecommandlockouts                              
\usepackage[pdftex]{graphicx}
\usepackage[cmex10]{amsmath}
\usepackage{amssymb}
\usepackage{import}
\usepackage{bm}

\usepackage{color}
\usepackage{units}
\usepackage{url}
\usepackage{float}
\usepackage{epstopdf}
\usepackage{stmaryrd}

\usepackage{units}
\usepackage{siunitx}

\usepackage{outlines}

\usepackage{boldline}
\usepackage[flushleft]{threeparttable}

\usepackage[doi=false,isbn=false,defernumbers=true,style=ieee,backend=bibtex,maxbibnames=1000]{biblatex}

\usepackage[utf8]{inputenc}
\bibliography{bibliography.bib}

\usepackage{mathtools}

\usepackage{color,soul}

\input{macros}

\title{\LARGE \bf
Optimal measurement selection algorithm and estimator for ultra-wideband symmetric ranging localization
}

\author{Saman Fahandezh-Saadi and Mark W. Mueller
\thanks{The authors are with the Department of Mechanical Engineering, at the University of California, Berkeley.
\newline        {\tt \{samanfahandej,mwm\}@berkeley.edu}}%
}

\begin{document}

\maketitle
\thispagestyle{empty}
\pagestyle{empty}

\begin{abstract}

A state estimator is derived for an agent with the ability to measure single ranges to fixed points in its environment, and equipped with an accelerometer and a rate gyroscope. The state estimator makes no agent-specific assumptions, and can be immediately applied to any rigid body with these sensors. Also, the state estimator doesn't use any trilateration-based method to calculate position from range measurements. As the considered system can only make a single range measurement at a time, we present a greedy optimization algorithm for selecting the best measurement. Experiments in an indoor testbed using an externally controlled multicopter demonstrate the efficacy of the algorithm, specifically showing an improvement over a na\"ive strategy of a fixed sequence of measurements. In separate experiments, the algorithm is also used in feedback control, to control the position of the multicopter.
\end{abstract}

\section{Introduction}
Low cost, flexible, and reliable localization technology is a key enabling technology for robotics. 
One of the main current limitations for the deployment of autonomous agents is the agents' ability to reliably and accurately determine their position. 
Various different sensing technologies exist for localization, ranging from purely self-contained on the agent to globally distributed satellite navigation. 

Different technologies represent a variety of different trade-offs, with varying requirements of computational power, electric power, precision, reliability, and accuracy. 
Perhaps the most widely used localization technology relies on satellites (e.g. GPS or Galileo) -- these systems work reliably and accurately when the agent has a clear line-of-sight to the satellites (typically, outdoors, and far from tall structures), some example robotic systems are \cite{cho2011wind, wendel2006integrated, hui1998unmanned, barczyk2012integration, jang2003longitudinal}.  
However, the reliance on a clear line of sight to the satellites is also the main drawback, leading to very poor (or nonexistent) localization in the presence of tall structures (e.g. in cities) or indoors. 

In research laboratories, a popular in-door alternative is to use optical motion capture equipment, which can yield extremely precise measurements (errors on order of millimetres) at high rates, but only over small volumes (on the order of 100$m^3$).
Such systems provide extremely rich data, but they are expensive, fragile, and are very constrained -- for examples of robotic systems relying on motion capture are \cite{how2008real, michael2010grasp, lupashin2014platform}. 

Another paradigm relies exclusively on sensors on the agent itself, for example cameras or laser range finders.
Here, the agent may fuse the measurements with other sensors, to simultaneously build a map of its environment, and localize within it (the SLAM problem), \cite{forster2013collaborative,nutzi2011fusion,scaramuzza2014vision,cadena2016past}. 
Such systems are attractive, since they are self-contained, but also require substantial computational power, heavy sensors/cameras -- this leads to large, heavy, complex, and energy-hungry agents.
Such systems may also be fragile, especially in environments with visual changes (e.g. smoke, changing light conditions and shadows, etc.).

Alternatively, radio- or audio-beacons  can be installed indoor, to build an indoor analogue of GPS, such as in \cite{priyantha2001cricket,krishnan2007uwb,segura2009mobile,prorok2010indoor,mueller2015fusing}.
Such systems may be created out of relatively low-cost components, provide high-quality localization over large areas, and do not impose particularly large restrictions on the individual agents using the localization system. 
However, they do require the installation of infrastructure, and as such are less flexible than vision-based systems. Ultra-wideband (UWB) radio ranging is an example of such localization system. 

Utilizing the UWB as range sensor for localization purposes is well-developed in the literature. Range-only SLAM is a precise way of localizing wireless sensor networks (WSN) node positions \cite{WSN_SLAM}. In \cite{UWB_SLAM2010}, the UWB ranging sensor is used for the range-only SLAM approach. In \cite{sensorSelection}, a method is described based on the Fisher information matrix of the Kalman filter which improves the target tracking accuracy of the wireless network. The method is related to our proposed method in this paper which selects sensors for future measurements. 

\begin{figure}[b]
    \centering
    \includegraphics{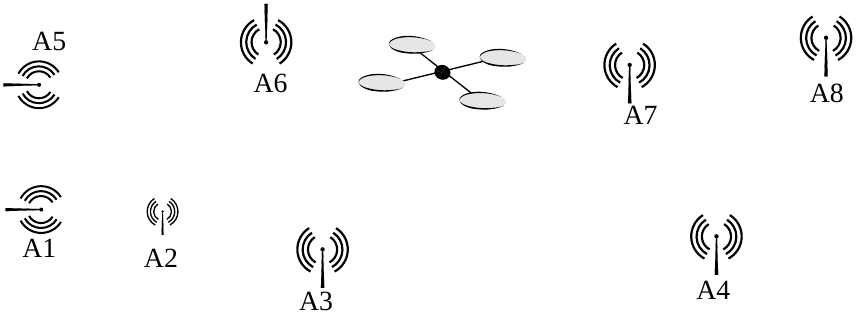}
    \caption{
    A schematic of the proposed systems: an agent (here a quadcopter) operates in a space prepared with multiple radio anchors at known locations (indexed A1-A8) in the Figure.
    At any given instant in time, the vehicle can only measure a distance to one anchor.
    }
    \label{fig:sysLayout}
\end{figure}

A schematic of UWB ranging system is given in \figref{fig:sysLayout}.
Furthermore, such systems typically use active measurements, wherein a measurement involves the transmission of a radio/audio message from the localization infrastructure to/from the agent.
Since these communications use the same frequency band, this imposes a constraint on the number of simultaneous measurements.

In this paper, we present a flexible state estimator, and an algorithm for selecting the optimal localization measurement, so that an agent may maximize its localization quality. 
A first-principles model is developed for an autonomous agent localizing by measuring distances to fixed (known) locations in the world, which is incorporated in a Kalman filter for six degrees-of-freedom (6DOF) state estimation. 
The results are validated in a series of experiments, where the estimator is deployed on a low-cost quadcopter system.

\section{Modelling}
\label{secModel}
We consider the problem of estimating the state of an agent, modelled as a generic 6 degree of freedom rigid body, equipped with an inertial measurement (accelerometer and rate gyroscope), and with the ability to measure the distance to any of a fixed set of points in its environment. 
The body's degrees of freedom are three in translation, and three in rotation; yielding a six-dimensional state vector to estimate. 
The goal is to have as general as possible a model, so that the resulting estimator may be applied to a variety of types of agents without modification.

\subsection{Equations of motion}
We will use the convention of using bold-face symbols for vector/matrix quantities, and regular font for scalars. 
Specifically, the position of the object is denoted as $\mvec{x}$, expressed in a coordinate system fixed with respect to the ground. 
The object's velocity and acceleration are given respectively by $\mvec{v}$ and $\mvec{a}$, again expressed in the ground. 
The orientation of the object is encoded with the rotation matrix $\mvec{R}$, and the angular velocity is given by $\mvec{\omega}$. 
The rotation matrix is defined so that multiplication by $\mvec{R}$ is equivalent to a coordinate transformation to the inertial frame, from the body-fixed frame. 

The time derivatives of these quantities are given as 
\begin{align}
   \ddt \mvec{x} &= \mvec{v} 
\\ \ddt \mvec{v} &= \mvec{a}
\\ \ddt \mvec{R} &= \mvec{R}\crossMat{\mvec{\omega}}
\end{align}
where $\crossMat{\mvec{\omega}}$ is the skew-symmetric matrix version of the cross product, so that $\crossMat{\mvec{x}}\mvec{y} = \mvec{x}\times\mvec{y}$.
Note we do not use the derivative of the angular velocity, as the angular velocity can be reliably estimated directly from the rate gyroscope outputs. 

\subsection{Inertial measurements}
The agent's inertial measurement unit outputs accelerometer and rate gyroscope measurements, $\mvec{\alpha}$ and $\mvec{\gamma}$, respectively. 
Both sensors are assumed to be well-calibrated, specifically having no scale errors nor bias offset. 

The rate gyroscope measures the angular velocity, corrupted by an additive noise $\mvec{\nu}_\gamma$:
\begin{align}
\mvec{\gamma} = \mvec{\omega} + \mvec{\nu}_\gamma
\end{align}
The accelerometer measures the `proper acceleration' of the vehicle, in the vehicle's body-fixed coordinate system, as given by:
\begin{align}
	\mvec{\alpha}_m = \mvec{R}^{-1} \mrb{\mvec{a} - \mvec{g}} + \mvec{\nu}_\alpha
\end{align}
where we again assume corruption by an additive noise $\mvec{\nu}_\alpha$, and $\mvec{g}$ is the gravitational acceleration vector, constant in the earth-fixed frame -- see \cite{leishman2014quadrotors} for a good tutorial.
A typical `z-up' coordinate system would have $\mvec{g}=\mrb{0,0,-9.81}\si{m\per s\squared}$.

\subsection{Range measurement system}
At discrete times, the agent is able to measure the distance from itself to one of a set of fixed positions in the world (here, called `anchors'). 
The anchors are at known positions $\mvec{p}_i$ in the world, and a measurement $\rho_i$ to anchor $i$ is modelled as
\begin{align}
    \rho_i = \mnorm{\mvec{x} - \mvec{p}_i} + \nu_\rho \label{eqModelRangeMeasurent}
\end{align}
where $\mnorm{\cdot}$ is the Euclidean norm, and $\nu_\rho$ is a scalar, additive noise.
Note that this assumes that the radio antenna is located at the same point as the inertial measurement unit; this assumption can be relaxed easily however.
Furthermore, no dependency is assumed on the orientation, though this has been shown to be a potentially important effect \cite{ledergerber2017ultra}. 

A single distance measurement between the agent and an anchor consists of a set of four radio messages, which allow the agent and the anchor to determine the distance between them by measuring the time-of-flight of the radio signal (see \cite{mueller2015fusing} for a similar scheme). 
The agent can communicate only with one anchor at a time, meaning that only a single range measurement can be taken at any instant in time. 

\section{Estimator}
\label{secEstimator}
The goal of the estimator is to estimate the 12-element state of the rigid body agent, using measurements from the inertial measurement unit and the range measurements. 
We create a ``kinematic'' state estimator for a generic 6DOF object, making specifically no assumptions on the forces or torques acting on the system.  
This yields a flexible estimator, that may be readily applied to a variety of rigid bodies; the flexibility comes at the cost of some precision (if we had an accurate model of the forces/torques acting on the agent, this information could be used to improve the estimator performance). 
The estimator is based on the Extended Kalman Filter (EKF) \cite{simon2006optimal}, specifically using the technique of \cite{mueller2016covariance} to encode an attitude in the state with correct-to-first-order statistics. It is worth mentioning that no trilateration method has been used to calculate the position of the agent from a set of range measurements. The estimator only relies on a single measurement at each step of EKF.

Although the 6DOF agent has twelve states, the estimator's stochastic state $\mvec{\xi}$ is 9 dimensional:
\begin{align}
    \hat{\mvec{\xi}} = \mrb{\hat{\mvec{x}}, \hat{\mvec{v}}, \hat{\mvec{\delta}}}
\end{align}
with the hat denoting estimated quantities, and where $\mvec{\delta}$ is an ``attitude error'' measure, assumed to be small.
The estimator uses a redundant attitude representation, with a `reference attitude' $\mvec{R}_\mathrm{ref}$ and the attitude error $\hat{\mvec{\delta}}$ combined yielding the estimator's attitude estimate $\hat{\mvec{R}}$
\begin{align}
    \hat{\mvec{R}} = \mvec{R}_\mathrm{ref} \mrb{I + \crossMat{\hat{\mvec{\delta}}}}
\end{align}
with $I$ the identity matrix.
This representation allows for a singularity-free attitude estimation using only a three-dimensional representation of the attitude error.
This is achieved by enforcing the requirement that $\mvec{\delta}$ is zero after each Kalman filtering step -- a complete discussion of this approach is given in \cite{mueller2016covariance}.

The estimator does not include the angular velocity $\mvec{\omega}$ as a state, and instead uses the measurement from the rate gyroscope directly.
This is justified by the high-quality measurements from modern rate gyroscopes, and is a standard approach in e.g. attitude estimation for satellites (see, e.g. \cite{markley2003attitude}).
Not only is this conceptually simpler than encoding additional states, it substantially reduces the computational complexity of the resulting state estimator (since the computational complexity of a Kalman filter scales approximately like the number of states cubed). 

In the prediction stage of the extended Kalman filter, the output of the accelerometer and rate gyroscope are used, so that the acceleration is given by:
\begin{align}
    \ddt \mvec{v} &= \mvec{R} \mvec{\alpha}_m + \mvec{g} + \mvec{R}\mvec{\nu}_\alpha
\end{align}
The orientation differential equation is rewritten in terms of $\mvec{\delta}$, with specifically
\begin{align}
    \ddt \mvec{\delta} &= \mvec{\gamma} - \mvec{\nu}_\gamma
\end{align}

It is assumed that the sensor noise terms $\mvec{\nu}_\alpha$ and $\mvec{\nu}_\gamma$ are zero-mean, and spatially and temporally independent, so that they can be straight-forwardly modelled as process noise in the Kalman filter formulation. 

The output from the ranging radios is used for the estimator's measurement update step. 
Specifically, given a measured distance $\rho_i$ (with the subscript $i$ indicating the choice of anchor to which the range was measured), the measurement equation \eqref{eqModelRangeMeasurent} can be linearized straight-forwardly to apply the Extended Kalman filter formulation. 

Since the measurement model is the distance of the agent from the anchor, the partial derivative with respect to the estimator state has an interesting property:
\begin{align}
    \mvec{H}_i :=& \frac{\partial \rho_i}{\partial \mvec{\xi}}  = \mrb{\frac{\partial \rho_i}{\partial \mvec{x}}, \frac{\partial \rho_i}{\partial \mvec{v}}, \frac{\partial \rho_i}{\partial \mvec{\delta}}} \label{eq:eqMeasPartialEqn}
\\  \frac{\partial \rho_i}{\partial \mvec{x}} =& \frac{\mvec{x}-\mvec{p}_i}{\mnorm{\mvec{x}-\mvec{p}_i}} =: \mvec{e}_i 
\\  \frac{\partial \rho_i}{\partial \mvec{v}} =& \frac{\partial \rho_i}{\partial \mvec{\delta}} = \mvec{0}
\end{align}
This means that the measurement sensitivity is the unit vector in the direction of the agent from the anchor $\mvec{e}_i$ -- a very intuitive property that will be exploited in the next section to determine which anchor $i$ should be used for the measurement. 

The Kalman filter also computes an estimated covariance matrix, $\mvec{\Sigma}$ relying on the partial derivatives and using the approach of \cite{mueller2016covariance}. 
The matrix may be partitioned into blocks, as below
\begin{align}
\matSym{\Sigma} = 
	\begin{bmatrix}	
		\matSym{\Sigma_{\mvec{x}\mvec{x}}} & \matSym{\Sigma_{\mvec{x}\mvec{v}}} & \matSym{\Sigma_{\mvec{x}\mvec{\delta}}}\\
		\matSym{\Sigma_{\mvec{v}\mvec{x}}} & \matSym{\Sigma_{\mvec{v}\mvec{v}}} & \matSym{\Sigma_{\mvec{v}\mvec{\delta}}}\\
		\matSym{\Sigma_{\mvec{\delta}\mvec{x}}} & \matSym{\Sigma_{\mvec{\delta}\mvec{v}}} & \matSym{\Sigma_{\mvec{\delta}\mvec{\delta}}}\\
	\end{bmatrix} \in \mathbb{R}^{9\times9}
\end{align} 

with e.g. $\mvec{\Sigma}_{\mvec{x}\mvec{\delta}}$ the $3\times3$ cross-covariance between the position and attitude states. 

\section{Anchor selection algorithm}
\label{secSelection}
As the agent is only capable of measuring the distance to a single anchor at any given time, there is the freedom to choose which anchor. 
A simple algorithm to use is to proceed sequentially through the list of anchors, consistently following some pre-determined ordering. Here, instead, we describe a computationally efficient and greedy selection algorithm that maximizes the information gain from the anchors at each time step. This is done, specifically, by choosing to get that measurement which produces the largest decrease in the estimator's variance, using the matrix trace as measure of size. This is closely related to the information matrix (Fisher information), which for a Gaussian case is the same as the inverse of covariance matrix, and provides the measure of information about the state present in the observations \cite{Mutambara:1998:DEC:522415}. This means by minimizing the covariance matrix, the maximum of available information in the measurements can be extracted.

Given a measurement $\rho_i$ from anchor $i$, the Kalman filter's covariance is updated at each time instant according to the standard Kalman filter equations: 
\begin{align}
    \mvec{K}_i &= \matSym{\Sigma}\mvec{H_i}^\transpose(\mvec{H_i}\matSym{\Sigma}\mvec{H_i}^\transpose + r)^{-1} \label{eq:kfGain}
	\\\matSym{\Sigma}^+_i &= \mrb{I - \mvec{K}_i\mvec{H_i}}\matSym{\Sigma} \label{eq:meas_update}
\end{align}
where $\mvec{\Sigma}_{i}^{+}$ is the updated covariance matrix after a measurement update, $\mvec{K}_i$ is the EKF gain matrix, and $\mvec{H_i}$ is the measurement matrix as computed in \eqref{eq:eqMeasPartialEqn}. By substituting \eqref{eq:kfGain} in \eqref{eq:meas_update}, we define the change in covariance due to a measurement update from anchor $i$ as $\Delta \mvec{\Sigma}_i$
\begin{align}
	\Delta \matSym{\Sigma}_i &= \matSym{\Sigma}^{+}_i - \matSym{\Sigma} = - \matSym{\Sigma} \mvec{H}_i^\transpose(\mvec{H}_i\matSym{\Sigma} \mvec{H}_i^\transpose + r)^{-1}\mvec{H}_i\matSym{\Sigma} \label{eq:var_diff}
\end{align}
with $r$ the (scalar) variance of the ranging noise $\nu_\rho$. Since the measurements are scalar, the matrix inverse is simply an ordinary division. Also, the matrix $H_i$ is sparse, therefore \eqref{eq:var_diff} can be decomposed as 
\begin{align}
    &\Delta \matSym{\Sigma}_i = \frac{-1}{\mvec{e}_i^\transpose\matSym{\Sigma_{\mvec{x}\mvec{x}}}\mvec{e}_i + r}
    \begin{bmatrix}
        \matSym{\Sigma_{\mvec{x}\mvec{x}}}\mvec{e}_i \\ 
        \matSym{\Sigma_{\mvec{x}\mvec{v}}}\mvec{e}_i \\ 
        \matSym{\Sigma_{\mvec{x}\mvec{R}}}\mvec{e}_i
    \end{bmatrix}
    \begin{bmatrix}
        \matSym{\Sigma_{\mvec{x}\mvec{x}}}\mvec{e}_i \\ 
        \matSym{\Sigma_{\mvec{x}\mvec{v}}}\mvec{e}_i \\ 
        \matSym{\Sigma_{\mvec{x}\mvec{R}}}\mvec{e}_i
    \end{bmatrix}^\transpose
    \label{eq:structured_eq3}
\end{align}
consisting of the projection of the variance onto the unit vector to the anchor $i$.

By comparing this change in covariance, different potential measurements (to anchors at different locations) at any given time can be compared. 
Notable is the intuitive form that this change takes, which will be exploited to generate an easily computed metric, below.

\subsection{Minimizing covariance matrix trace}
The performance metric we minimize is the trace of the covariance matrix after the measurement, that is $\tr\mrb{\mvec{\Sigma}^+_i}$. The trace of covariance matrix is the mean of the norm of the error squared of state estimator (i.e. $\tr\mrb{\mvec{\Sigma}^+_i} = \E ||\hat{\mvec{\xi}}-\mvec{\xi}||^2$). This metric perfectly makes sense, since the goal is to reduce the estimation error as much as possible \cite{Boyd:2004:CO:993483}.

From the definition of $\Delta \mvec{\Sigma}_i$ in \eqref{eq:structured_eq3}, and the linearity of the trace operator, it follows that this is equivalent to maximizing $\tr\mrb{\Delta \mvec{\Sigma}_i}$ (i.e. by maximizing the difference between covariance of Kalman filter before and after measurement, the algorithm chooses the anchor which reduces the covariance trace the most).

Starting from equation \eqref{eq:var_diff}, using the cyclic property of matrix traces (e.g. $\tr(\matSym{AB}) = \tr(\matSym{B}\matSym{A})$), we simplify to get
\begin{align}\label{eq:wo_trace}
    \tr(\Delta \matSym{\Sigma}_i) 
    & = -\frac{\mvec{H}_i\matSym{\Sigma}\matSym{\Sigma}\mvec{H}_i^\transpose}{\mvec{H}_i\matSym{\Sigma} \mvec{H}_i^\transpose + r}
\end{align}
This can be simplified further by exploiting the sparsity of $\mvec{H}_i$, to yield
\begin{align}\label{eq:each_anc}
    \tr(\Delta \matSym{\Sigma}_i) = - 
    \frac{\mnorm{\matSym{\Sigma_{\mvec{x}\mvec{x}}} \mvec{e}_i}^2 + \mnorm{\matSym{\Sigma_{\mvec{x}\mvec{v}}} \mvec{e}_i}^2 + \mnorm{\matSym{\Sigma_{\mvec{x}\mvec{\delta}}} \mvec{e}_i}^2} 
    {\mvec{e}_i^\transpose\matSym{\Sigma_{\mvec{x}\mvec{x}}}\mvec{e}_i + r}
\end{align}
At a given instant in time, the system computes $\Delta \mvec{\Sigma}_i$ for each anchor $i$ in the network, and then selects the maximizing anchor. This is a simple computation, requiring only a small number of multiplications to compute, and therefore easily implemented on computationally limited hardware. (i.e. using \eqref{eq:each_anc} with the necessary number of fixed points in the space (usually 5 or 6), the computation cost is significantly small considering relatively powerful microcontroller used in today's robotic systems).

\section{Experimental validation}
\label{secExpValidation}
\begin{figure*}
    \centering
    \makebox[0pt]{\includegraphics{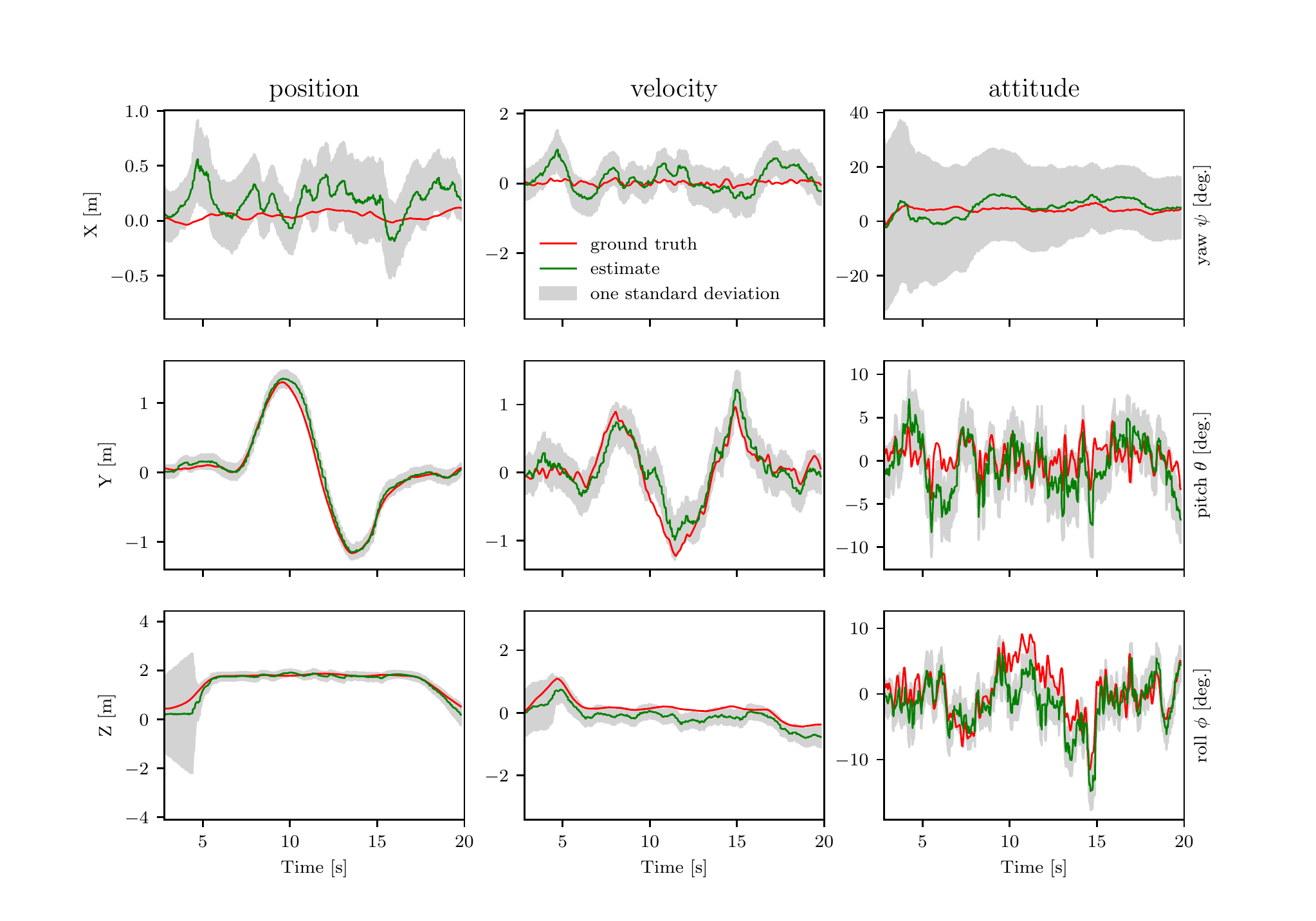}}
    \caption{
    {Estimation using ultra-wideband ranging localization (optimal anchor selection).} The left column of plots shows the position in $[\si{m}]$, the middle column shows the velocity in $[\si{m/s}]$, and the right column shows the attitude of the quadcopter in [deg.]. The experiment illustrates the comparison of true states (driven from motion caption system) against our estimator using the optimal anchor choice. As seen, quadcopter takes off and hovers for $2\si{s}$, then starts moving along the $y$ axis horizontally for $8 sec.$, and finally lands at the origin. The yellow area around the mean values on the plots shows the square roots of the diagonals of EKF covariance matrix (i.e. one standard deviation).
    Note that the estimator does not compute the estimate in terms of yaw, pitch, and roll angles; the estimator output is simply transformed into this format as it is easier to parse in a figure.
    }
    \label{fig:openloop}
\end{figure*}
The approach is validated in experiment, where a quadcopter is used as the autonomous agent.
A first set of experiments is an ensemble, showing the performance of the algorithm for the quadcopter under external control, so that the motion is repeatable.
These experiments compare performance when using the greedy optimization to that when using a fixed, sequential measurement sequence. 
The second experiment demonstrates closed-loop control of the quadcopter, using the resulting state estimate for feedback control. 

\subsection{Experimental Setup}
\begin{figure}
    \centering
    \includegraphics[width=0.8\linewidth]{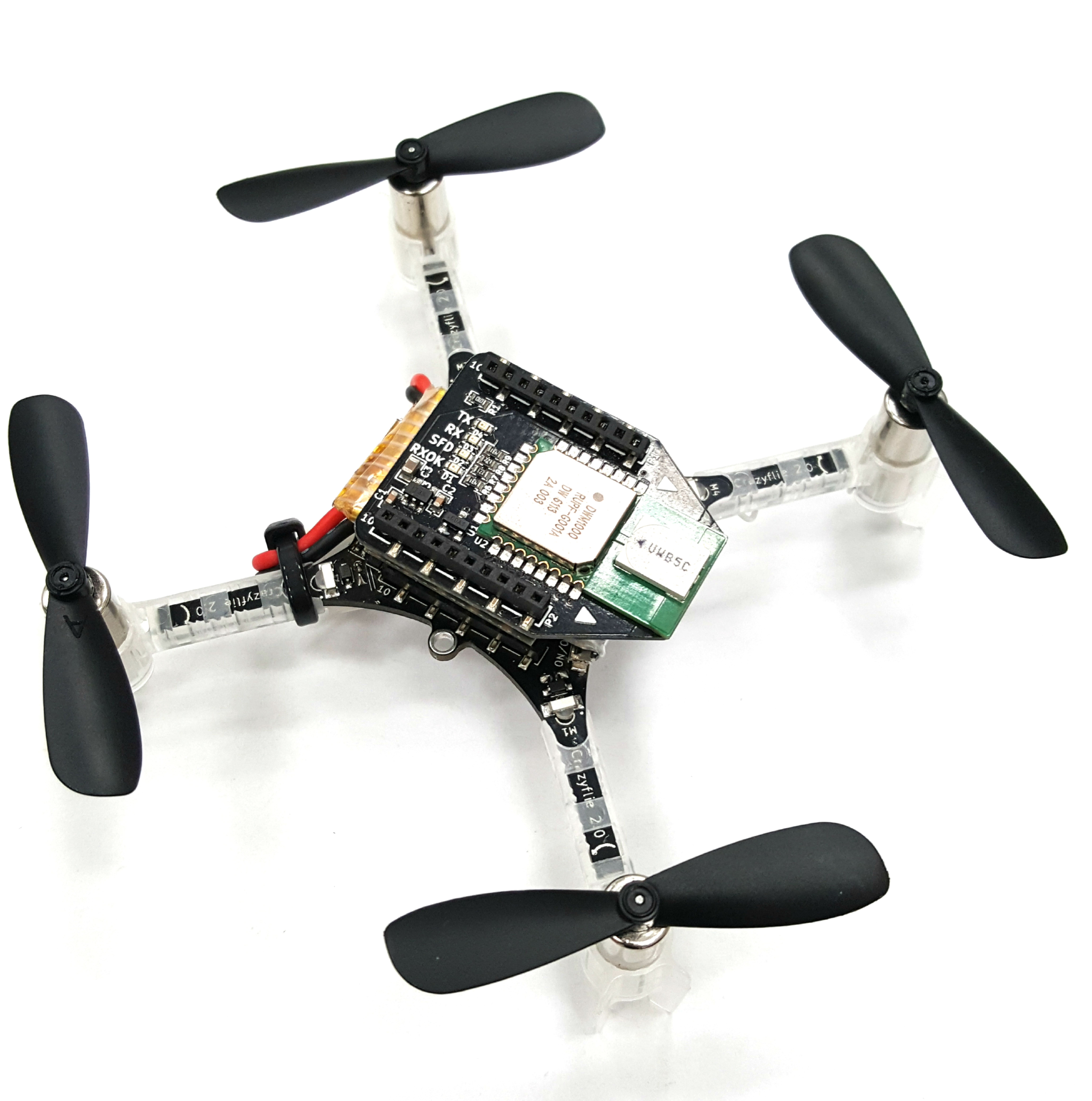}
    \caption{
    The quadcopter used in our experiments as the mobile agent. The UWB ranging sensor (sees in this picture) works at the rate of approximately 60 Hz.
    } 
    \label{fig:cf}
\end{figure}

We tested our algorithm on a Crazyflie 2.0 quadcopter (shown in \figref{fig:cf}), with approximate mass of $30 \si{g}$, and a scale of approximately $105 \si{mm}$.
The quadcopter is equipped with an STM32F4 microcontroller, uses an Invensense MPU9250 inertial measurement unit, and a Decawave DW1000 module for the ultra-wideband ranging measurements. 
The anchors shared the same computational and sensing hardware. 
All computations for the state estimation (including the measurement selection) were performed on the microcontroller. 
Measurements from the accelerometer and rate gyroscope were taken at $500\si{Hz}$, and range measurements were taken at approximately $60\si{Hz}$. 
The estimator performance is quantified by using a ceiling-mounted motion capture system, whose measurements are taken as ground truth. 

\begin{figure}
    \centering
    \makebox[0pt]{\includegraphics[scale = 0.45]{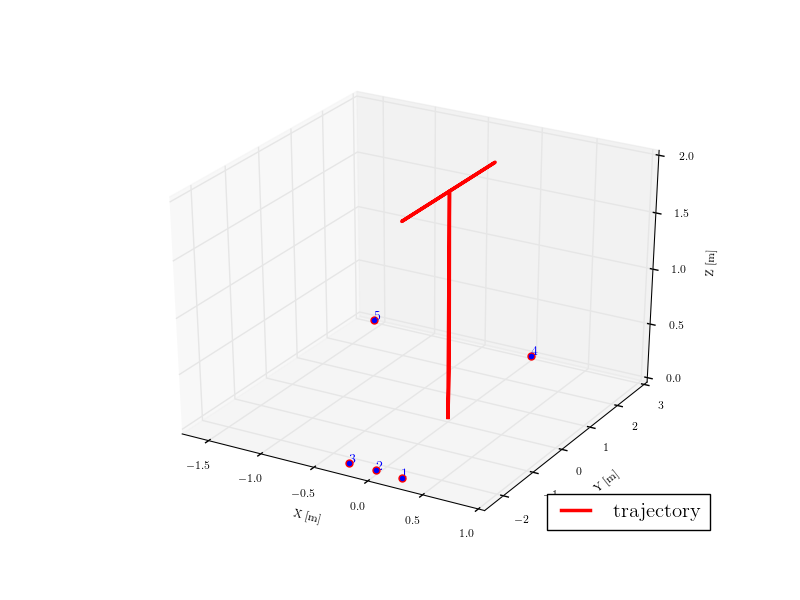}}
    \caption{Anchor setup and quadcopter trajectory. The set of five anchors are all located at $z=0$, and the quadcopter's motion is confined to a plane where $x=0$, with the horizontal motion at a height of $2\si{m}$.
    }
    \label{fig:anchors}
\end{figure}

The anchor arrangement as well as the quadcopter's commanded trajectory can be seen in \figref{fig:anchors}. In general at least four anchors is needed in order to estimate a unique position in the space (otherwise the estimate of position may be arbitrarily rotated, reflected, or even translated). But in real-time implementation, it is a good practice to have more than four anchors for redundancy. Since the estimator uses only one range measurement at a time from one anchor, increasing the number of anchors will not affect the results theoretically, though this should be investigated in real-time experiment. Notable is that three anchors are placed in very close proximity to one another, so that their measurements convey very similar information. This was chosen so as to highlight the effect of the greedy optimization algorithm. 
\subsection{Estimation only}
In the first set of experiments, the state estimate is not used for control, and instead an offboard controller is used.
This offboard controller uses measurements from the motion capture system, and consists of a set of cascaded controllers similar to that of \cite{lupashin2014platform}.
The use of the offboard controller resulted in repeatable experiments, where the quadcopter moved along very similar trajectories for each flight (which therefore yields a fair comparison). For all experiments, the estimator of \secref{secEstimator} runs exclusively on the agent's microcontroller. 

We examined and compared the estimator outputs with two different EKF settings: when the quadcopter sequentially ranges to anchors (the na\"ive approach), and when the quadcopter uses the optimization of \secref{secSelection}. The reference trajectory path is also shown in \figref{fig:anchors}.

\begin{table}
\begin{threeparttable}
\centering
\caption{Comparison of estimation error from experiments}
\label{tab:rmse}
\begin{tabular}{|l|l|l V{} l|l V{} l|l|}
\hline
 & \multicolumn{6}{c|}{RMSE} \\
\hline
trials & \multicolumn{2}{c V{}}{position [m]} & \multicolumn{2}{c V{}}{velocity [m/s]} & \multicolumn{2}{c|}{attitude [deg.]} \\
\hline
{} & seq. & opt.\tnote{a}& seq. & opt. & seq. & opt.\\
\hline
1 & 0.286 & 0.274 & 0.638 & 0.536 & 8.1 & 5.3\\
2 & 0.278 & 0.267 & 0.440 & 0.432 & 6.9 & 4.8\\
3 & 0.312 & 0.291 & 0.635 & 0.585 & 8.9 & 5.7\\
4 & 0.236 & 0.245 & 0.415 & 0.474 & 7.9 & 7.5\\
5 & 0.489 & 0.363 & 0.875 & 0.731 & 7.6 & 7.9\\
6 & 0.347 & 0.273 & 0.589 & 0.525 & 6.7 & 8.6\\
7 & 0.310 & 0.263 & 0.585 & 0.489 & 8.7 & 6.7\\
8 & 0.436 & 0.383 & 0.821 & 0.690 & 14.0 & 7.6\\
9 & 0.228 & 0.262 & 0.436 & 0.426 & 6.4 & 6.4\\
10 & 0.422 & 0.329 & 0.866 & 0.593 & 8.3 & 8.2\\
\hline
Avg. & 0.334 & 0.295 & 0.630 & 0.548 & 8.4 & 6.9\\
\hline\hline
diff. & \multicolumn{2}{c V{}}{-11.7\%} & \multicolumn{2}{c V{}}{-13.0\%} & \multicolumn{2}{c V{}}{-17.5\%}\\
\hline
\end{tabular}
        \begin{tablenotes}
            \item[a] seq. for sequential ranging, opt. for optimal anchor selection, as defined in \secref{secSelection}
        \end{tablenotes}
\end{threeparttable}
\end{table}

The experiment was repeated ten times for each algorithm, and the resulting root mean square error (RMSE) for each trial are shown in Table \ref{tab:rmse}. 
Due to the system's stochastic nature, the best run using the na\"ive sequential selection approach is better than the worst run using the optimization algorithm.
Nonetheless, a clear improvement is observed on average, with an RMSE reduction of approximately 11\% in position and velocity, and 17\% in attitude when using the optimal measurement selection algorithm.

\figref{fig:openloop} shows experimental data from a representative trial using the optimal anchor selection algorithm. 
The graph shows that the estimated state is close to the ground truth data from the motion capture system, with specifically also the estimator variance being a reliable indication of estimate quality. The results are closely related to our conclusion on maximizing the most informative direction in the space. The extreme anchor arrangement, gives very little information in the $x$ direction to the estimator, and this creates large uncertainty on that direction. Notable also is the large initial uncertainty in attitude for the rotation about gravity. This is because this state is unobservable without horizontal motion, and the uncertainty rapidly decreases when the vehicle executes the side-to-side motion.  

\subsection{Closed-loop Tracking Experiment}
\begin{figure}
    \begin{center}
    \makebox[0pt]{\includegraphics{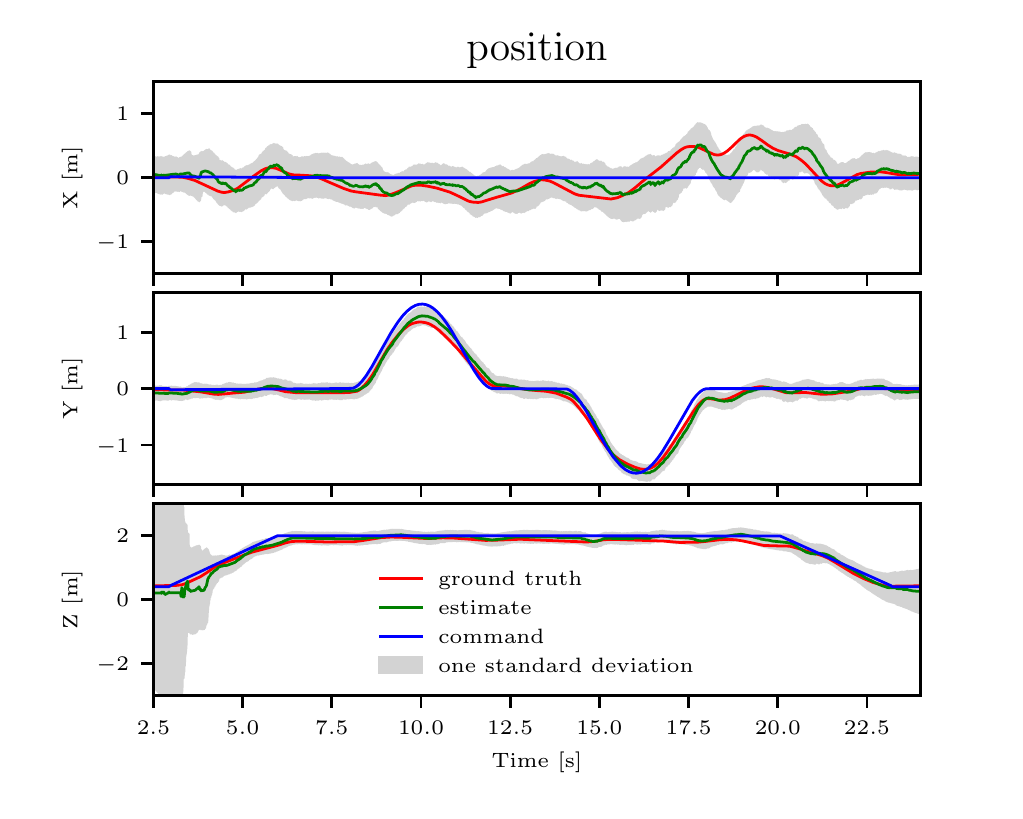}}
    \caption{
    Closed-loop control of the quadcopter using the presented optimal anchor selection algorithm.
    The plot represents the position of quadcopter. 
    The blue, green, and red lines show the command position, the onboard position estimator (EKF), and the motion capture system output (ground truth), respectively. 
    The yellow area around the estimation shows one standard deviation on position in each direction.
    }
    \label{fig:closedloop}
    \end{center}
\end{figure}

The second experiment demonstrates that the estimator performs sufficiently well to allow the quadcopter to fly without the use of the external system. 
The result of the experiment in position can be seen in \figref{fig:closedloop}. 
As seen in the graph, the quadcopter tracks the command in $y$ and $z$ directions well, but performs more poorly in the $x$ direction. 
This is due to the larger uncertainty in this direction, stemming from the anchor layout. 

\section{Conclusion}
\label{secConclusion}
The paper presented a computationally low-cost algorithm for estimating the six degree of freedom state of a rigid body equipped with an inertial measurement unit and operating inside a system of range-measuring radios.
Specific attention was paid to the optimal selection of a ranging measurement, where the resulting approach is based on greedily minimizing the trace of the estimator covariance matrix. Although increasing the rate of measurement can potentially improve the estimator performance, the hardware limitation is always an issue. Also, even with higher measurement rate, the proposed algorithm can be used to improve the performance further more. The resulting algorithm can be deployed on low-cost, computationally constrained devices. The estimator has the advantage that it makes no specific assumptions about the agent's motion, and could thus be immediately applied to a large variety of agents.

A series of flight experiments were performed to demonstrate the efficacy of the estimator as implemented on a low-cost quadcopter. 
Over twenty repeated experiments with the quadcopter externally controlled, a substantial improvement due to the selection of the anchor using the greedy optimization approach compared to using a na\"ive sequential ranging algorithm was shown. 
Further experiments showed that the estimator performs sufficiently well to be used for closed-loop control as well, even for fast, unstable systems such as quadcopters.

\section{Acknowledgement}
\label{secAcknowledgement}
This material is based upon work supported by the National Science Foundation Graduate Research Fellowship under Grant No. DGE 1106400.

\section*{References}
{
\printbibliography[heading=none, resetnumbers=true]
}

\end{document}

%% file: macros.tex
\newcommand{\mnorm}[1]{\left\| #1 \right\|}

\newcommand{\mvec}[1]{\boldsymbol{#1}}

\newcommand{\matSym}[1]{\boldsymbol{#1}}

\newcommand{\mrb}[1]{\left( #1 \right)} 

\DeclareMathOperator*{\E}{\mathbb{E}} 

\newcommand{\mnoshow}[1]{}

\newcommand{\figref}[1]{Fig.~\ref{#1}}
\newcommand{\secref}[1]{Section~\ref{#1}}

\newcommand{\crossMat}[1]{\ensuremath{\llbracket{#1}\rrbracket}}

\newcommand{\ddt}{\frac{\mathrm{d}}{\mathrm{d}t}}

\newcommand*{\transpose}{T}

\newcommand{\tr}{\mathrm{tr}}